\documentclass[review]{elsarticle}

\usepackage{color,xcolor}
\usepackage{lineno,hyperref}
\usepackage{textcomp}
\usepackage{graphics}
\graphicspath{}
\modulolinenumbers[5]
\biboptions{sort&compress}
\usepackage{amsmath}
\usepackage{verbatim}
\usepackage{cleveref}
\usepackage{bbding}
\usepackage{hyperref}
\usepackage{amssymb}
\usepackage{latexsym}
\usepackage{graphicx}
\usepackage{subfigure}

\usepackage{algorithm}  
\usepackage{algpseudocode}  
\usepackage{amsmath}  
\usepackage{caption} 
\usepackage{array}
\usepackage{txfonts} 
\usepackage{ragged2e}
\usepackage{url}

\usepackage{array} 

\usepackage{booktabs} 

\usepackage{longtable}

\usepackage{bm}

\usepackage{setspace} 

\usepackage{natbib} 

\begin{document}

\captionsetup[figure]{labelfont={bf},name={Fig.},labelsep=period}
\captionsetup[table]{labelfont={bf},name={Table.},labelsep=period}
\begin{frontmatter}
	
	\title{Dynamic Image Restoration and Fusion Based on Dynamic Degradation}
	\author{Aiqing Fang, Xinbo Zhao$^*$, Jiaqi Yang, Yanning Zhang}%
	\address{
		\justifying\let\raggedright\justifying
		$^a$National Engineering Laboratory for Integrated Aero-Space-Ground-Ocean Big Data Application Technology, School of Computer Science, Northwestern Polytechnical University, Xi’an 710072, China
		
	}
	
	
	
	\cortext[mycorrespondingauthor]{Corresponding author}
	\ead{xbozhao@nwpu.edu.cn (X. Zhao), aiqingf@mail.nwpu.edu.cn (A. Fang),
	jqyang@nwpu.edu.cn (J. Yang), caoshihao@mail.nwpu.edu.cn (S. Cao),	
	ynzhang@nwpu.edu.cn (Y. Zhang)}
	
	\begin{abstract}
The deep-learning-based image restoration and fusion methods have achieved remarkable results. However, the existing restoration and fusion methods paid limited research attention to the robustness problem caused by dynamic degradation. In this paper, we propose a novel dynamic image restoration and fusion neural network, termed as DDRF-Net, which is capable of solving two problems, i.e., static restoration and fusion, dynamic degradation. In order to solve the static fusion problem of existing methods, dynamic convolution is introduced to learn dynamic restoration and fusion weights. In addition, a dynamic degradation kernel is proposed to improve the robustness of image restoration and fusion. Our network framework can effectively combine image degradation with image fusion tasks, provides more detailed information for image fusion tasks through image restoration loss, and optimizes image restoration tasks through image fusion loss. Therefore, the stumbling blocks of deep learning in image fusion, e.g., static fusion weight and specifically designed network architecture, are greatly mitigated. Extensive experiments show that our method neatly outperforms the state-of-the-art methods.
	\end{abstract}
	
\end{frontmatter}

\section{Introduction}
As a basic research content in the field of computer vision, image fusion plays an important role in many high-level semantic tasks, e.g., object detction, saliency detection and object recognition. The existing image fusion methods have done a lot of exploration on the fusion criteria (e.g., maximum, sum, weighted average, and $L1$) and network framework (e.g., Siamese network, nest network, and generative adversarial networks). However, existing image fusion methods lack exploration and analysis of dynamic fusion weight and dynamic degradation model. Unfortunately, in real-world scenes, due to the differences between data acquisition equipment and environment, different cross-modal images are faced with dynamic degradation factors, making the existing static image fusion weights unable to adapt to the dynamic scene. 

However, the human vision system can handle the dynamic scene robustly, which are closely related to the dynamic cognitive processing mechanism of human brain.  According to the research of cognitive psychology \cite{HUTCHISON2013360} and biological neuroscience \cite{2015Dynamic}, the cognitive process of the human brain is dynamic. The dynamic characteristics of human brain for information collection, transmission and processing make human beings complete very complex visual and auditory tasks. \textit{Therefore, we believe that the dynamic cognitive learning of the human brain is of positive significance to improve the robustness of image fusion tasks, as verified in Sect.4.}

\begin{figure}[ht]
\centering
\includegraphics[width=1.0\textwidth]{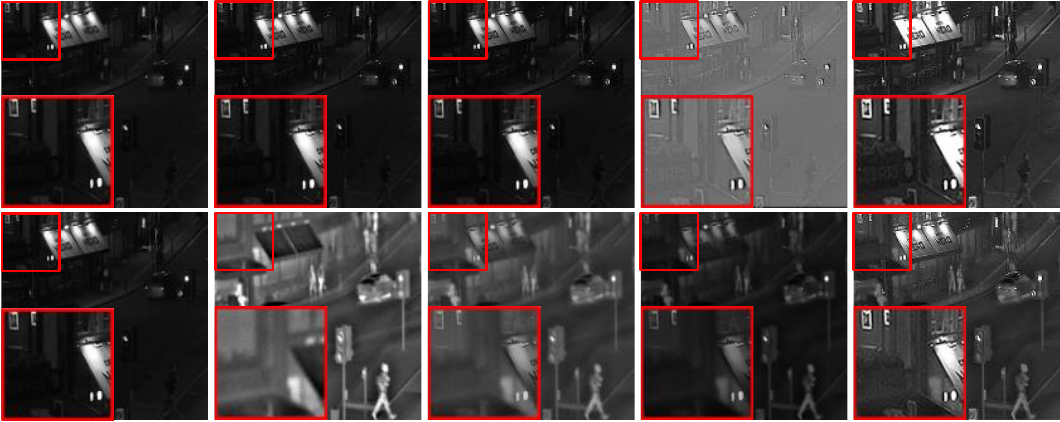}
\caption{Examples of image restoration and fusion. The first line image is an example of image restoration, and the second line image is an example of image fusion. The first two columns of images are used as the input of the network, and the middle image is the existing image restoration and fusion methods. The last column of images is our method.\label{summary}}
\end{figure}

In order to overcome dynamic image fusion problem, inspired by the human brain dynamic cognitive mechanism, a robust dynamic image restore and fusion method is proposed. \textit{As far as we know, our research is the first exploration in the field of image fusion and image restoration.} Our method studies dynamic fusion weights and dynamic noise kernels in kernel space. Because this method processes data in kernel space, it is more efficient than methods leveraging feature space. Different from the existing image restore and fusion methods, the input of our feature extraction module is not only the cross-modal images but also the dynamic degradation kernel. This method can extract the features of different modal images and restore the degraded images at the same time. This joint optimization strategy is critical to hidden feature extraction. In addition, dynamic convolution is used to construct the image fusion module, which makes the fusion weight follow the sample with dynamic characteristics. Moreover, we introduce the dynamic negative sample fusion loss through the idea of adversarial learning, so as to reduce the image fusion noise and improve the quality of image fusion.
The main contributions of our work include the following four points:
\begin{itemize}

\item \textbf{Firstly}, we make a research on the influence of static fusion weight and static noise on image fusion tasks and propose the idea of dynamic fusion, i.e. dynamic fusion weight, dynamic noise kernel, and dynamic negative sample fusion loss.

\item \textbf{Secondly}, we propose a dynamic image fusion network with the ability of image restoration. Through the joint optimization of image fusion and image restoration tasks, the robustness of image fusion in a complex environment (i.e., low light, and blur) is further improved.

\item \textbf{Finally}, we make comprehensive comparative and analysis experiments on image restoration and fusion on public datasets. The experimental results demonstrate our method is superior compared with the state-of-the-art methods.	

\end{itemize}
The remainder of this paper is structured as follows. Sect. 2 reviews relevant theory knowledge. Sect. 3 presents a dynamic image fusion method. Sect. 4 introduces the experimental datasets, evaluation metrics, and implementation details. Sect. 5 presents a discussion and explanation. Sect. 6 draws a conclusion.

\section{Related work}
Our research involves not only dynamic image fusion but also dynamic image degradation. Therefore, we briefly introduce image restoration, dynamic convolution and image fusion.
\subsection{Dynamic convolution}
Deep convolution neural network \cite{2021Dynamic} has achieved remarkable results in many visual fields. However, the traditional convolution module often obtains static learning weights, which will not change in the test phase once the training is completed. This will severely limit the performance of the network model in changing scenarios. To solve this problem, the concept of dynamic convolution (e.g., CondConv \cite{NIPS2019_8412}, DynamicConv \cite{Chen_2020_CVPR}, and DyNet \cite{zhang2020dynet}) is proposed, i.e., the target convolution kernel is obtained by weighting a series of convolution kernels. Dynamic convolution has excellent performance in the task of classification and recognition but there is a lack of relevant research in the task of image fusion.
\subsection{Image restoration}
When existing image restoration \cite{Zero-DCE} or super-resolution \cite{zhang2018learning, leiz} tasks generate degraded data, they often use limited degradation kernel (e.g., blurring, bicubicly downsampling, and white Gaussian noise) to simulate image degradation in a real environment. However, in this way, a single image will only be affected by a single blur kernel and downsampling, which is inconsistent with the real environment, i.e., a single image will be affected by a variety of degradation kernels. 
Image degradation is of great research value not only for image restoration task or image super-resolution task, but also for image fusion task, especially for cross-modal image fusion task without ground truth. After all, image fusion is to improve the real quality of the image, not the similarity of original images. Unfortunately, no relevant research has been found in the field of cross-modal image fusion.

\subsection{Image fusion}
Due to the outstanding performance of deep learning methods in many visual fields, deep-learning-based image fusion methods have been widely proposed,e.g., DenseFuse \cite{Li2018DenseFuse}, IFCNN \cite{ZHANG202099}, FusionDN \cite{xu2020aaai}, PGMI \cite{PGMI}, U2Fusion \cite{U2Fusion}, NestFuse \cite{li2020nestfuse}, RFN-Nest \cite{LI202172}. Existing deep-learning-based image fusion networks, e.g., image fusion methods based on Siamese networks and generative adversarial networks , often use similarity loss (e.g., content loss, detail loss, gradient loss, and tee loss) or adversarial loss to learn static fusion weight. For the cross-modal image fusion task without ground truth, it is necessary to calculate the similarity between the predicted image and the original image, so as to retain the texture details of source images as much as possible. Li et al. \cite{Li2018DenseFuse} introduced a SSIM \cite{1284395} to optimize the image fusion weight. Many cross-modal image fusion methods (e.g., IFCNN, FusionDN, NestFuse, and U2Fusion) based on deep learning use this similarity measure optimization function, e.g., SSIM, MSE, and content loss. However, Ma et al. \cite{MaFusionGAN} introduced an adversarial loss into the infrared and visible image fusion task. Although the introduction of adversarial loss provides a new idea for image fusion, the subjective fusion quality obtained by adversarial loss has an obvious local blur effect. Therefore, to solve this problem, Ma et al. \cite{9031751} introduced a dual discrimination generative adversarial fusion network that combined multiple loss functions (e.g., content loss, detail loss, gradient loss, tee loss, and adversarial loss) are used to optimize the fusion weight. In addition, the fusion criterion of existing image fusion methods is usually defined as the fusion layer after the feature extraction layer, which separates the fusion from the subsequent decoding convolution. Although this method is widely used, there are also some irrationality. 1) The fusion criterion of artificial design needs specific expert experience. 2) Artificially designed fusion standards will limit the versatility of the network, such as infrared and visible light, multi-focus image fusion tasks. To solve above problems, we carried out the related research of dynamic image fusion and proposed the dynamic degradation for image restoration and fusion.

\section{Proposed method}
\begin{figure}[h]
	\centering
	\includegraphics[width=0.9\textwidth]{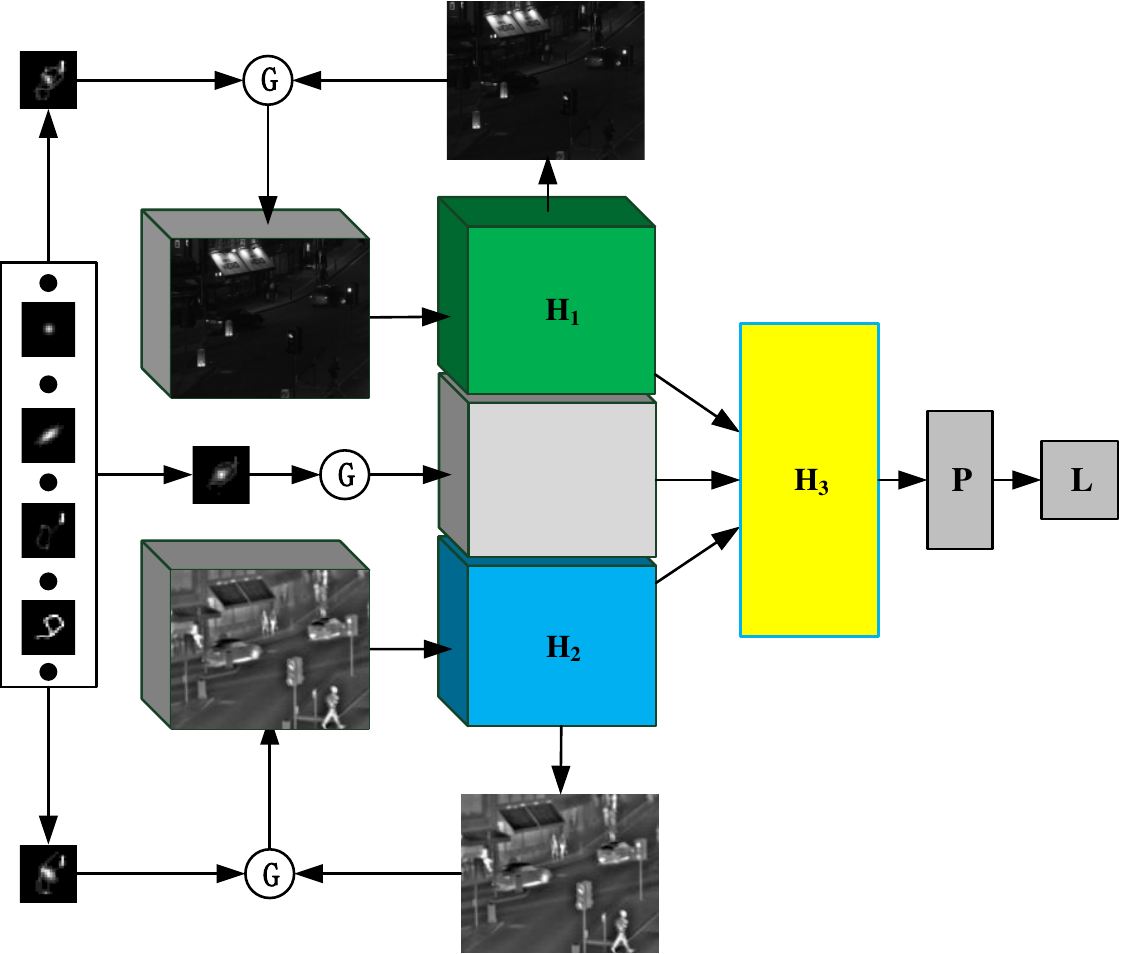}
	\caption{Network Architecture of the proposed DDRF-Net, where $H_1$ represents the dynamic degradation model of visible image; $H_2$ represents the dynamic degradation model of infrared image; $H_3$ represents the dynamic degradation model of fused image, i.e., fusion process; $L$ represents the loss of image fusion task; $P$ denotes the fused image; $G$ denotes PCA and stretch alignment operation \cite{zhang2018learning}. The dynamic convolution kernel is composed of 12 basic kernels of 4 types (i.e., isotropic Gaussian kernel, motion blur kernel, and anisotropic Gaussian kernel) in the black box.\label{framework}}
\end{figure}
In this section, with analysis of the static restoration and fusion methods, we provide our dynamic restoration and fusion formulation, the definition and design of dynamic kernel. At last, the neural network architecture design and training skills are shown concretely.

\subsection{Motivation}
The robustness of image fusion is of great significance for its application and development. The existing image fusion methods lack research on dynamic fusion weight and dynamic degradation model, i.e., the visual data is affected by illumination, blur, and compression transmission at the same time, which is dynamic and coupled with each other, as verified in section \ref{setup}. The existing image fusion methods do not consider this dynamic degradation problem in the feature extraction stage. In the dynamic visual scene, the original feature extraction weight will be invalid, and the image fusion effect will be seriously affected. Although there is comprehensive research on the degenerate kernels in the existing image restoration tasks, only the influence of a single degenerate kernel (e.g., downsampling combined with blur kernel) on the overall visual scene is considered. However, dynamic fusion weight and dynamic degradation models play an important role in improving the robustness of image fusion. Therefore, we are motivated to study dynamic fusion and dynamic degradation models in the field of image fusion.

\subsection{Dynamic degradation}
Image restoration and fusion are not only to calculate the similarity between the predicted image and different modal images but also to restore high-quality image. Although existing image restoration methods have done a lot of research on different degradation kernels, existing degradation kernels are mostly static degradation kernels, only considering the global degradation of the image (i.e., a single degradation kernel is used for the whole image), without considering the local degradation of the image (i.e., a single image will be affected by different degradation models). The traditional degradation model H on Z (high quality image) is defined as \cite{leiz}:

\begin{equation}
 {Z \circledast H}=(\bm {Z} \circledast \bm  {k}) \downarrow_{s}+ {N},
\end{equation}
where $\bm k$ indicates a 2-D convolution kernel, denotes convolution operation, ↓s represents down-sampling process with a scaling factor s, N indicates white gaussian noise. The existing degradation convolution $\bm k$ is a single static convolution kernel, which cannot simulate the degradation of real images. Therefore, in order to better simulate the real degradation model, we define $\bm k$ as a linear combination of a series of basic degradation kernels, i.e., isotropic Gaussian kernel, motion blur kernel, and anisotropic Gaussian kernel. The dynamic kernel is defined as:
\begin{equation}
\begin{array}{r}
 {k}_{d}= {a} *  {k}_{m}(\bm {j})+ {b} * {k}_{i}(\bm {j})+ {c} * {k}_{a}(\bm {j})\\
\text { s.t. }
\mathrm{j}=\{1,2,3,4\} \& {a}+{b}+{c}= {1} \& \mathrm{a}, \mathrm{b}, \mathrm{c} \in [0,1]
\end{array},
\end{equation}
where a, b, c indicate the weight of basic degradation kernel, $k_m(\bm j)$ denotes motion blur kernel, $k_i(\bm j)$ indicates isotropic Gaussian kernel, $k_a(\bm j)$ indicate anisotropic Gaussian kernel, $x$ respresents a random integer variable. In order to avoid the image oscillation caused by the dynamic degradation kernel, the weights $a, b, c$ must be added to sum to 1. Therefore, the dynamic degradation model $H_m$ can be formulated as:
\begin{equation}
{Z}_i *  {H}_m=\left(\bm {Z}_i \circledast \bm {k}_d\right) \downarrow_{s}+ {N},  {m} \in \left\{  1,   2,   3\right\}
\end{equation}

For the cross-modal image fusion task, taking the infrared and visible image fusion task as an example, we need to restore three dynamic degradation models, i.e., visible image $(H_1)$, infrared image $(H_2)$, and fused image $(H_3)$.

\subsection{Dynamic fusion}
Existing image fusion methods often use artificial fusion criteria in feature space or learn fixed fusion weights by the data-driven method. 
The fusion criterion for artificial design is defined as:
\begin{equation}
 {ø}=\sum_{i=1}^{k}\left( {w}_i(\bm {x}, \bm {y}) *   {C}\left(\bm {f}_{i }, \bm{f}_{v }\right)\right),
\end{equation}
where $\bm f_i$ and $\bm f_v$ denote infrared and visible feature maps; $w_i(\bm x, \bm y)$ denotes the fusion weight of the position of feature map $ (\bm x, \bm y)$; $ø$ indicates the fused feature maps; $C(*)$ denotes concatenate operation; $ø$ indicates fused feature maps. This method focuses on the design and improvement of the fusion weight strategy of different feature maps, e.g., the fusion weight based on saliency, weighted average, maximum, and sum. For static CNN, the fusion weight can be formulated as:
\begin{equation}
 {ø}={  {W^T} *  {C}\left(\bm {f}_{i }, \bm {f}_{v }\right) }+{\bm {b}},
\end{equation}
where $W^T$ and $b$ denote the convolution process. To overcome the shortcomings of the existing fusion methods (i.e., artificial fusion criterion and static fusion weight), we propose a dynamic fusion method in kernel space. We concatenate the convolution kernel parameters learned by the feature extraction module, and linearly stack the convolution kernel weights combined with the channel attention mechanism. The dynamic fusion criteria is defined as:
\begin{equation}
\mathfrak{ø}=\left( (\frac{1}{N} \sum_{\mathrm{k}=1}^{N}\pi_{k}(\boldsymbol{x}) {\boldsymbol{W}}_{k})^T*\mathbf{C}\left(\bm {f}_{i}, \bm {f}_{v}, \bm k_{di}, \bm k_{dv} \right)+ \frac{1}{N} \sum_{\mathrm{k}=1}^{N}\pi_{k}(\boldsymbol{x}) {\boldsymbol{b}}_{k}\right),
\end{equation}
where $W_k$ denotes k-th convolution kernel; $N$ the number of convolution kernel of dynamic kernel; $\pi_{k}(\boldsymbol{x})$ denotes convolution kernel weight, i.e., channel attention weight; $C(\bm f_i, \bm f_v, \bm k_{di}, \bm k_{dv})$ denotes the dynamic degenerate kernel (i.e., $\bm k_{di}$, and $\bm k_{dv}$) is superimposed on the feature maps, i.e., visible feature map ($\bm f_v$), and infrared feature map ($\bm f_i$) in the channel dimension.

\subsection{Loss Function}
Existing deep-learning-based image fusion methods mainly use similarity measure loss to optimize image fusion weight, and the similarity fusion loss ${\cal L(\bm \theta)}$ is defined as:

\begin{equation}
{\cal L(\bm \theta)}= 1 - \frac{1}{N}\sum\limits_{i=1}^{N} S(f(\bf X_i,\bm \theta),P),
\label{gs50}
\end{equation}
where $\bm {X_i}$ indicates i-th image to be fused; $\bm \theta$ denotes CNN's weight; $\bm P$ indicates fused image; $S(*)$ denotes similarity function. For the cross-modal image fusion task, due to the limitations (i.e., just represents similarity, rather than high-quality image) of the existing similarity loss, this paper introduces the idea of adversarial learning based on the dynamic degradation model and proposes the dynamic negative sample fusion loss ${\cal L(\bm H_i)}$ that is defined as: 

\begin{equation}
{\cal L(\bm H_{i})}= 1 - \frac{1}{N}\sum\limits_{i=1}^{N} S(\bf X_i* \bm H_i,P),
\label{gs51}
\end{equation}
where $\bm H_i$ denotes i-th dynamic model. The basis of adversarial learning is to have positive samples as guidance so that the network model can be optimized in the right direction. Therefore, we introduce a positive sample loss ${\cal L(\bm \theta)}$, which is defined as:
\begin{align}
{\cal L(\bm O)} &= O(\bf{ M_1, M_2, M_3, \cdots, M_i})
\end{align}
where $M_i$ denotes i-th image fusion method; $O(*)$ indicates evaluation function. Therefore our fusion loss function L is defined as:
\begin{equation}
{\cal L(\bm \theta)}= \frac{1}{3}\sum\limits_{i=1}^{3} {\cal L(\bm H_{i})}+ \cal L(\bm \theta)+ \cal L(\bm O)
\label{gs52}
\end{equation}
Our fusion loss considers not only the similarity with the original image, but also the negative sample distance based on the dynamic degradation model. This method can effectively reduce the universality of image fusion caused by the difference of data distribution.

\subsection{Network architecture and training}
Our network framework adopts the classic Siamese network framework. In the two-way feature extraction branch network, the network weights are not shared. Traditional image fusion methods based on Siamese network or self-coding network, e.g., Deepfuse, Densefuse, and IFCNN, two sub-modules are purely for extracting features from different images. Different from the traditional image fusion method based on Siamese networks or self-coding network, our feature extraction branch is not only to extract features but also to restore the dynamic degradation model. Notably, the most important thing is that our network does not need datasets from other fields to assist learning. i.e., the cross-modal image fusion task completes the restoration of the dynamic degradation model with its own data. This strategy effectively reduces the complexity of training and avoids the problem of model recovery caused by the difference in data distribution. Different from the existing deep-learning-based methods (e.g., NestFuse, RFN-Nest, and SAF), our network does not need complex network structure, e.g., RESNET, NestNet, and DenseNet. Our fusion module uses the superposition of four layers of dynamic convolution, and the weight of the convolution kernel is generated by the channel attention mechanism. In the process of training, our dynamic degradation kernel will be serialized, dimension reduced and stretched to align with different modal data in dimension. When the same image is affected by multiple degradation kernels, it is difficult for a simple single branch network to converge. That is reason why we avoid a data-driven training model in feature space. 

\section{Experiments}
\label{setup}
In this section, experimental setup are first presented. Then, comparative experiment and analysis experiment are introduced.

\begin{table*}[!h]
	
	\centering \scriptsize
	\renewcommand \arraystretch{1.1}
	\caption{Metrics}
	\label{table123}
	\begin{tabular}[b]{p{0.05cm}p{1.3cm}p{4cm}p{5cm}}
		\hline
		No.& Method &Equations   & Description \\ 
		\hline
		1&EN \cite{1576816}&$EN=-\sum_{i=0}^{255} p_{i} \log _{2} p_{i},$&where $P_i$ is the probability of a gray level appearing in the image.\\ 
		\hline
		2&AG \cite{Cui2015Detail}&$AG=\frac{1}{M^{*} N} \sum_{i=1}^{M} \sum_{j=1}^{N} \sqrt{\frac{\Delta I_{x}^{2}(\bm i,\bm j)+\Delta I_{y}^{2}(\bm i,\bm j)}{2}},$& where $M \times N$ denotes the image height and width; $\Delta I_{\bm x}(\bm i,\bm j)$ denotes image horizontal gradient; $\Delta I_{y}(\bm i,\bm j)$ denotes image vertical gradient.\\ 
		\hline
		3&SSIM \cite{1284395}& $\begin{array}{l}
		{SSIM}(\bm I_i, \bm R)=\frac{\left(2 u_{I_i} u_{R}+C_{1}\right)\left(2 \sigma_{I_i R}+C_{2}\right)}{\left(u_{I_i}^{2}+u_{R}^{2}+C_{1}\right)\left(\sigma_{I_i}^{2}+\sigma_{R}^{2}+C_{2}\right)},\end{array}$&where $\mu_{I_i}$ and $\mu_R$ indicate the mean value of origin image $I_i$ and fused image $R$; $\sigma_{I_iR}$ is the standard covariance correlation.\\ 
		\hline
		4&VIFF \cite{Han2013A}& $\mathrm{VIF}=\frac{\sum_{j \in  { subbands }} I\left({C} \stackrel{N, j}{;} {F}^{N, j} | s^{N, j}\right)}{\sum_{j \in \ { subbands }} I\left({C}^{N, j} ; {E}^{N, j} | s^{N, j}\right)},$& where ${C} \stackrel{N, j}{;}$ denotes N elements of the $C_j$ that describes the coefficients from subband j; $\sum_{j \in  { subbands }} I\left({C} \stackrel{N, j}{;} {F}^{N, j} | s^{N, j}\right)$ denotes reference image information.\\  
		
		\hline	
		5&PSNR \cite{SijbersJ1996Qaio} &$\mathrm{PSNR}=10 \log _{10} \frac{\left(2^{n}-1\right)^{2}}{M S E},$& where $MSE$ is the mean square error of the current image X and the reference image y.\\ 
		
		\hline
	\end{tabular}
\end{table*}

\begin{table*}[!h]
	
	\centering \footnotesize
	\renewcommand \arraystretch{1.3}
	\caption{Parameter settings of evaluated methods. Bold values indicate different architectures of OURS. Time was computed on images with an average size of 400x400. The numbers in brackets indicate the size of the pre-trained model. }
	\label{table112}
	\scriptsize
	\begin{tabular}[b]{p{0.1cm}p{1.6cm}p{3cm}p{0.3cm}p{1.4cm}p{0.6cm}p{0.6cm}p{1cm}}
		\hline \\
		\textbf{No.}& \textbf{Method} &\textbf{Parameters} &\textbf{Year}  & \textbf{Category}  &\textbf{Time(s)}&\textbf{Dynamic }&\textbf{Model Size(M)} \\ \hline 
		1&IFCNN \cite{ZHANG202099}&$L_{r0}=0.01, power=0.9$&2020&Deep learning & {0.021}&$\times$ &$0.32(170)$\\
		
		2&GANMCC \cite{ma2021ganmcc}&$epoach=10, c_dim=1, scale=3, stride=14$	&2020&GAN& $\times$&$\times$&$10.90$\\ 
		
		3&PGMI \cite{PGMI} &$Epoach=15, lr=1e-4, c_dim=1, stride=14, scale=3$&$2020$&Deep learning&0.044 &$\times$&$0.80$\\
		4&NestFuse-max \cite{li2020nestfuse} &$Nb_{filter}=[64,112,160,208,256]$&$2020$&Deep learning &$0.116$ &$\times$&$20.80$\\
		
		5&U2Fusion \cite{U2Fusion} &$num=30, epoach=[3,2,2], lam=0$&$2020$&Deep learning &0.857&$\checkmark$&1064.96\\
		
		6&GFF \cite{MA2020103016}&$f=1, mode=1, n=10$ &2020&Traditional&0.094&$\times$&$\times$\\ 
		
		7&RFN-Nest \cite{LI202172} &$batchsize=4, epoch=2, \lambda = 100$&2021&Deep learning &0.119&$\checkmark$&18.20\\
		
		8&DDRF-Net & $batchsize=16, lr=0.001, k_b = 12$&2021& Deep learning&0.028&\checkmark&7.58\\
		\hline
	\end{tabular}
\end{table*}

\subsection{Experimental Setup}
In this section, datasets, metrics and methods for experimental evaluation are first presented. Then, implementation details of evaluated methods are introduced.

\textit{1) Datasets: We carry out experiments on four datasets, i.e., FLIR, RGB-t, gun, and TNO.}

\textit{a) FLIR} \cite{FLIR}: FLIR dataset is obtained by RGB and thermal imaging camera installed on the vehicle and contains 14, 452 thermal infrared images, including 10, 228 from short video and 4, 224 from the 144-second video. Unfortunately, there is no registration. 

\textit{b) RGB-T} \cite{LiChenglong2019RotB}: The RGBT dataset includes 821 image pairs.

\textit{c) Gun} \cite{ZHANG202099, Zhou2015Perceptual}: Gun is captured in night mode contaminated with blur and noise. There is a lot of duplication between this dataset and the TNO dataset. Therefore, we only use one image of $gun A \& gun B$ in this dataset.

\textit{d) TNO \cite{TNO}}: It contains multi-spectral (near-infrared and long-wave infrared or thermal) night images of different military-related scenes, registered in different multi-band camera systems. There are 21 pairs of images commonly used in existing image fusion methods.

\textit{2) Metrics:} The distinctiveness of an image quality is usually quantitatively evaluated using entropy (EN) \cite{1576816}, average gradient (AG) \cite{Cui2015Detail}, structural similarity (SSIM) \cite{1284395}, visual information fidelity (VIF) \cite{Han2013A}, natural image quality evaluator  (NIQE) \cite{MittalA2013MaCB}, PSNR \cite{SijbersJ1996Qaio}, are defined as \ref{table123}.

\textit{3) Methods:} As shown in Table \ref{table112}, we compared 7 state-of-the-art methods, i.e., 5 deep-learning-based methods, 1 GAN method, and 1 traditional method, published in the top journal.

\textit{4) Implementation details:} 
Before the experiment, we need to clarify the following questions. All the experimental data were tested in the same environment. Our experimental platform is desktop 3.0 GHZ i5-8500, RTX2070, 32G memory. 

\begin{figure}[h]
	\centering
	\includegraphics[scale=0.5,width=0.85\textwidth]{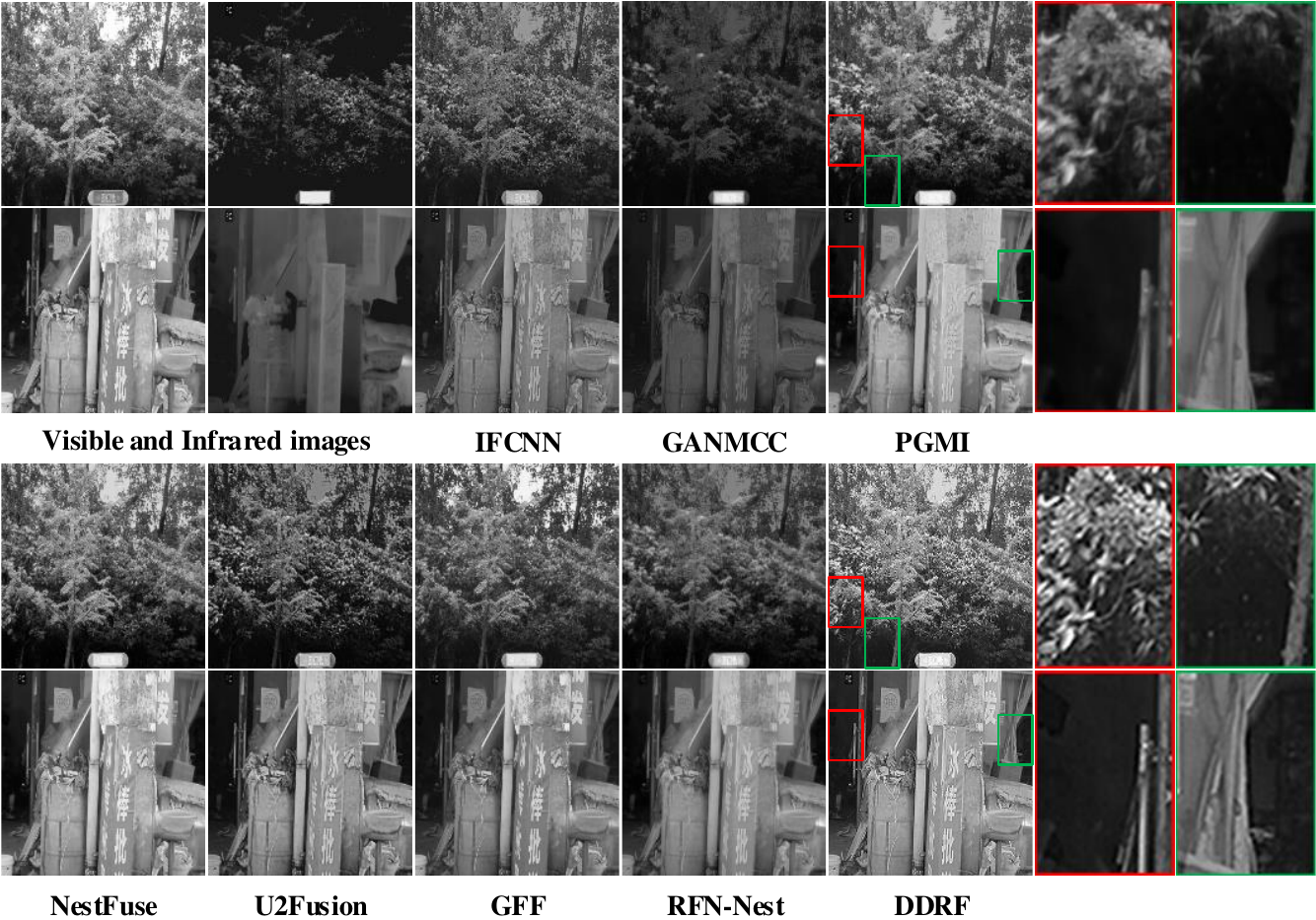}
	\caption{Examples of tested methods on the RGBT dataset with low-light and blur degradation.\label{RGBT}}
\end{figure}

\subsection{Comparative experiments}
In order to highlight the advantages of this method, we carried out comparative experiments and analytical experiments on the task of cross-modal image fusion and image restoration.

\subsubsection{Image Fusion}
\begin{figure}[t]
	\centering
	\includegraphics[width=0.85\textwidth]{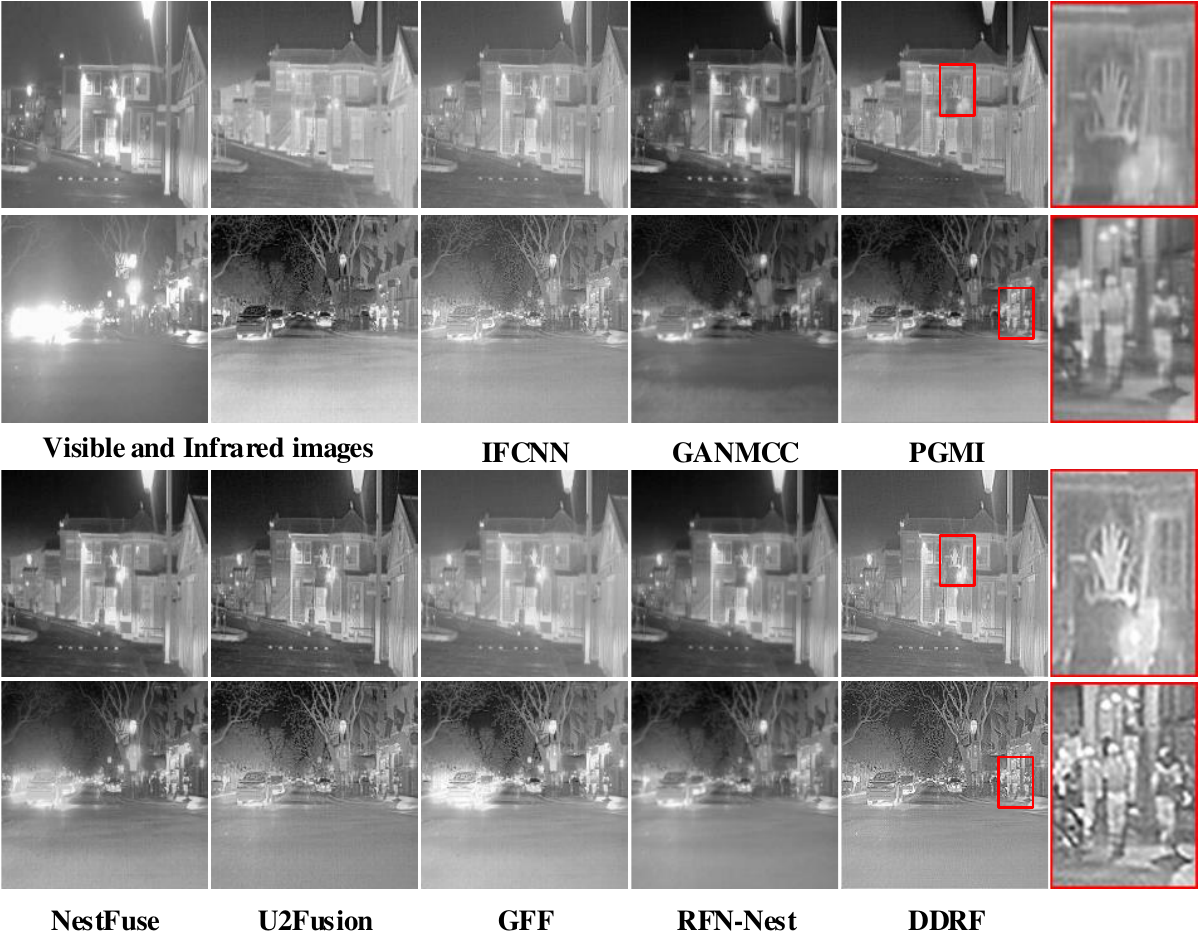}
	\caption{Examples of tested methods on the FLIR dataset with high-light, blur and noise degradation.\label{FLIR}}
\end{figure}

In order to show the superiority of our method, we test DDRF-Net and state-of-the-art methods on RGBT and FLIR datasets. In Figure \ref{RGBT}, there are a lot of low-light and blur problems in the visible images, only our fusion method still has good performance in these two cases. In Figure \ref{FLIR}, there are a lot of highlights and halos in visible image, and there are blur and noise in infrared image. We can see that IFCNN, U2Fusion and NestFuse have good results in high-light. The NestFuse is relatively poor in halo and blur effect processing, the RFN-Nest method is not good in blur and highlight processing. However, our method achieves the best results in highlight, halo and blur processing compared with state-of-the-art methods. From Figure \ref{FLIR} and Table \ref{f8}, we can see that our method has a very significant improvement in gradient compared with other methods, which means that our method has better definition.

\begin{table}[h]
	\centering \scriptsize
	\caption{Six evaluation indicators for quantitative contrast of tested methods on FLIR and RGBT datasets. Red color represents the largest objective index, and the green color represents the second.}
	\label{f8}
	\scalebox{1.1}{
		\renewcommand\tabcolsep{3pt} 
		\renewcommand\arraystretch{1.3}
		\begin{tabular}{lllllll|llllll}
			\hline
			Datasets &  &  &  &FLIR  &  &  &  &  &  RGBT&  &  &  \\ \hline
			Metrics & EN & AG & SSIM & VIF & PSNR & Mean & EN & AG & SSIM & VIF & PSNR & Mean \\ \hline
			DDRF & 7.22  & \textcolor{red}{9.43}  & 0.70  & \textcolor{green}{0.38}  & \textcolor{green}{43.66}  & \textcolor{red}{10.59}  & \textcolor{red}{7.71}  & \textcolor{red}{21.02}  & 0.58  & 0.31  & 35.33  &\textcolor{red}{ 11.23}  \\ 
			U2Fusion & 7.41  & \textcolor{green}{6.77}  & 0.70  & \textcolor{red}{0.40}  & 36.53  & 9.05  & 7.41  & 12.98  & 0.61  & 0.32  & 35.12  & 9.85  \\ 
			RFN & \textcolor{green}{7.46 } & 2.79  & 0.69  & 0.28  & 34.63  & 7.95  & 7.34  & 6.54  & 0.60  & 0.29  & 35.21  & 8.83  \\ 
			PGMI & 7.37  & 4.66  & 0.70  & 0.34  & 36.74  & 8.73  & \textcolor{green}{7.54}  & 9.71  & \textcolor{green}{0.62}  & 0.28  & 35.37  & 9.32  \\ 
			NestFuse & \textcolor{red}{7.60}  & 4.79  & 0.68  & \textcolor{green}{0.38 } & 35.02  & 8.56  & \textcolor{green}{7.71}  & 13.07  &\textcolor{green}{ 0.62}  & \textcolor{red}{0.41}  & \textcolor{red}{39.79}  & \textcolor{green}{11.08}  \\ 
			IFCNN & 7.09  & 5.92  & \textcolor{red}{0.76}  & 0.36  & \textcolor{red}{44.56}  & \textcolor{green}{10.19}  & 7.24  & \textcolor{green}{14.11}  & \textcolor{red}{0.64}  & 0.32  & 34.19 & 9.89  \\ 
			GFF & 7.34  & 5.06  & \textcolor{green}{0.75}  & 0.37  & 43.65  & 10.08  & 7.50  & 12.85  & \textcolor{red}{0.64}  & \textcolor{green}{0.34}  & 35.41  & 10.03  \\ 
			GANMCC & 7.32  & 3.90  & 0.61  & 0.30  & 28.66  & 7.14  & 6.51  & 5.33  & 0.53  & 0.24  & \textcolor{green}{35.65}  & 8.43  \\ \hline
	\end{tabular}}
\end{table}

\begin{figure}[t]
	\centering
	\includegraphics[width=0.9\textwidth]{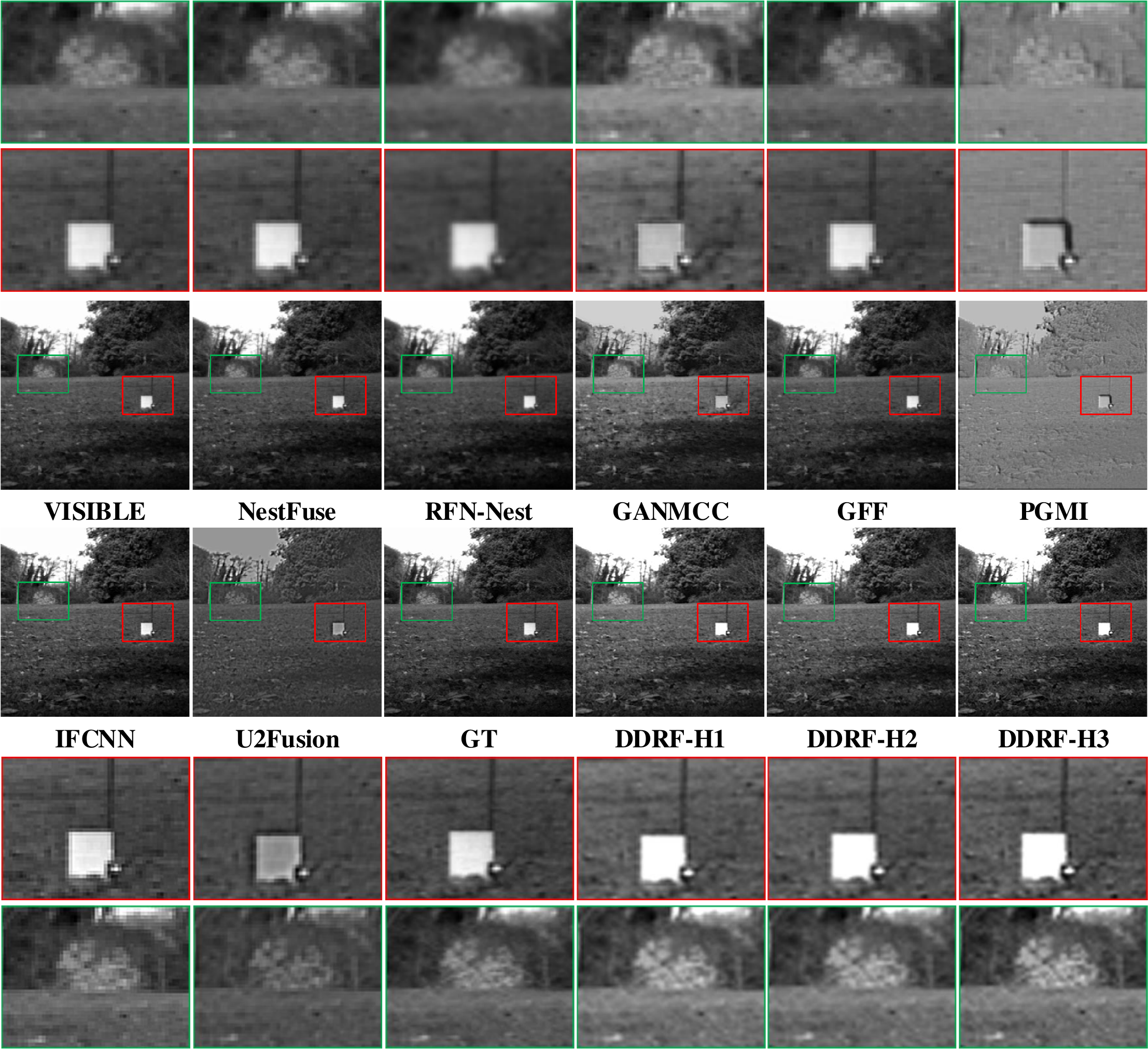}
	\caption{Examples of tested methods on the Gun dataset with noise and down-sampling degradation.\label{restore}}
\end{figure}

\subsubsection{Image restoration}
 \begin{figure}[h]
	\centering
	\includegraphics[width=0.9\textwidth]{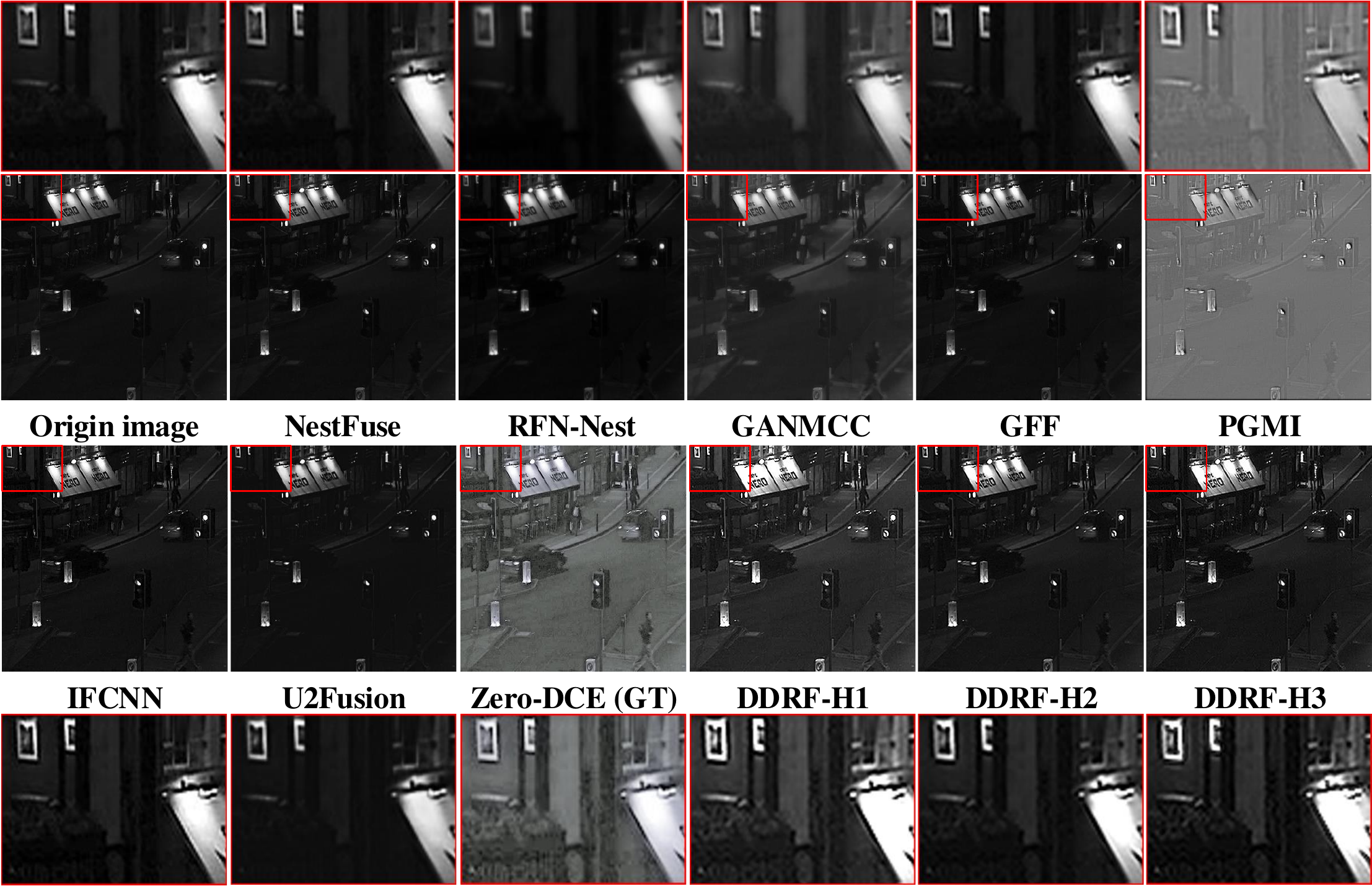}
	\caption{Examples of tested methods on the TNO dataset with low-light degradation\label{restore1}}
\end{figure}

In this experiment, for the dynamic image restoration task, we only need to convert the cross-modal image data into the same input data in the test phase. Our dynamic recovery model includes pure dynamic recovery, (i.e., $H_1, H_2$) and fusion dynamic recovery, i.e., $H_3$. In the test dataset of the Gun, for the problem of downsampling change, we downsampling the visible light image, so this experiment has ground truth labels (GT). For the problem of illumination change, we use the TNO dataset. However, the GT labels cannot be obtained by illumination change. Therefore, to solve the problem of brightness change, we use the special image restoration method i.e., Zero-DCE \cite{ Zero-DCE} published in CVPR in 2020, as ground truth (GT). Notably, the most important thing is that Zero-DCE is only effective for brightness changes, and it cannot effectively restore the image quality for downsampling. That's why it cannot be used as a GT for downsampling experimental. From Figure \ref{restore}, \ref{restore1} and Table \ref{f9}, we find that DDRF-Net is not only effective for cross-modal image fusion tasks, but also for image restoration tasks. Among them, the simple image restoration task (i.e., $H_1, H_2$) has a better effect than the dynamic fusion restoration task (i.e., $H_3$). DDRF-Net surpasses the tested methods (e.g., U2Fusion, PGMI, NestFuse, and RFN-Nest) in restoration ability. Although the IFCNN has a good effect on the recovery ability, there is still a gap compared with the DDRF-Net. From the image effect, we can still find that the special image restoration algorithm Zero-DCE has achieved good results in brightness and noise but also lost a lot of texture details, and the contrast information is significantly reduced compared with the original image. In addition, we can also find that due to the difference of data distribution between infrared image and visible image, the performance of $H_1$ and $H_2$ in different data distribution is different. Notably, from Figure \ref{restore1} and Table \label{f9}, we also find that when the original image have edge pixel problem caused by down-sampling degradation. In this condition, compared with the existing methods,  $H_1$, $H_2$ achieved better results, i.e., effectively overcome the edge pixel problem. However, the fusion effect of $H_3$ is better than that of $H_1$ and $H_2$, i.e., the fusion result is clearer and has better subjective visual perception.

\subsection{Analysis Experiments}
In this section, we analyze the impact of static and dynamic on image restoration and image fusion tasks.

\subsubsection{Static and dynamic restore and fusion analysis}

Although comparative experiments show that DDRF-Net has better fusion performance than state-of-the-art methods, we still need to further analyze whether the dynamic model has better performance than the static model. In this experiment, we compared two tasks, i.e., static restoration and dynamic restoration, static fusion and dynamic fusion. All the training parameters of the network model are the same, i.e., epoach=89, lr=0.001, batchsize=16. In the static and dynamic restoration tasks, we use the traditional convolution and dynamic convolution as the basic network unit. The experiment consists of three parts, i.e., restor$e_0$ analysis, restor$e_1$ analysis, and fusion analysis. Restor$e_0$ analysis is a special image restoration optimization. Restor$e_ 1 $analysis is a joint optimized image restoration, which is used to test the difference between special image restoration optimization and joint optimization. Fusion analysis is used to compare the differences between static fusion and dynamic fusion.

\begin{table}[h]
	\centering \scriptsize
	\caption{Six evaluation indicators for quantitative contrast of tested methods on Gun and TNO datasets. Red color represents the largest objective index, and the green color represents the second.}
	\label{f9}
	\scalebox{1.1}{
		\renewcommand\tabcolsep{3pt} 
		\renewcommand\arraystretch{1.3}
		\begin{tabular}{lllllll|llllll}
			\hline
			Datasets &  &  &  &Gun  &  &  &  &  & TNO &  &  &  \\ \hline
			Metrics & EN & AG & SSIM & VIF & PSNR & Mean & EN & AG & SSIM & VIF & PSNR & Mean \\ \hline
			U2Fusion & \textcolor{red}{7.23}  & \textcolor{red}{7.87}  & 0.10  & 0.01  & 20.44  & 7.13  & 3.96  & 2.81  & 0.44  & 0.27  & 35.73  & 8.64  \\ 
			RFN & 6.85  & 3.00  & 0.61  & 0.28  & 46.56  & 11.46  & 4.88  & 1.74  & 0.29  & 0.24  & 33.16  & 8.06  \\ 
			PGMI & 5.98  & 5.80  & 0.42  & 0.17  & 27.02  & 7.88  & 5.16  & 4.86  & 0.67  & 0.38  & 31.15  & 8.44  \\ 
			NestFuse & 6.84  & 4.41  & 0.65  & 0.35  & 49.73  & 12.40  & 5.07  & 3.17  & 0.46  & 0.39  & 36.11  & 9.04  \\ 
			IFCNN & 6.93  & \textcolor{green}{7.82}  & 0.65  & 0.38  & 46.36  & 12.43  & 5.43  & 5.14  & 0.56  & 0.51  & 37.69  & 9.87  \\ 
			GFF & 6.86  & 4.35  & 0.65  & 0.35  & 50.02  & 12.45  & 5.09  & 3.22  & 0.48  & 0.39  & 36.55  & 9.15  \\ 
			GANMCC & \textcolor{green}{7.07}  & 5.48  & 0.62  & 0.29  & 37.91  & 10.27  & 5.63  & 2.88  & \textcolor{green}{0.77}  & 0.34  & \textcolor{red}{49.21}  & \textcolor{green}{11.76 } \\ 
			DDRF-H1 & 6.91  & 6.53  & \textcolor{green}{0.93}  & \textcolor{red}{0.77}  & 53.34  & 13.69  & \textcolor{red}{6.07}  & \textcolor{red}{6.67}  & \textcolor{red}{0.90}  & \textcolor{red}{0.59}  & \textcolor{green}{48.74 } & \textcolor{red}{12.59}  \\ 
			DDRF-H2 & 6.89  & 6.18  & \textcolor{red}{0.95}  & \textcolor{green}{0.76}  & \textcolor{red}{57.77}  & \textcolor{red}{14.51}  & 5.45  & 4.18  & 0.67  & 0.46  & 40.69  & 10.29  \\ 
			DDRF-H3 & 6.89  & 6.18  & \textcolor{red}{0.95}  & \textcolor{green}{0.76}  & \textcolor{green}{56.81}  & \textcolor{green}{14.32}  & \textcolor{green}{5.79}  & \textcolor{green}{6.42}  & 0.68  & \textcolor{green}{0.52}  & 40.38  & 10.76  \\ \hline
	\end{tabular}}
\end{table}

\begin{figure}[h]
	\centering
	\includegraphics[width=0.95\textwidth]{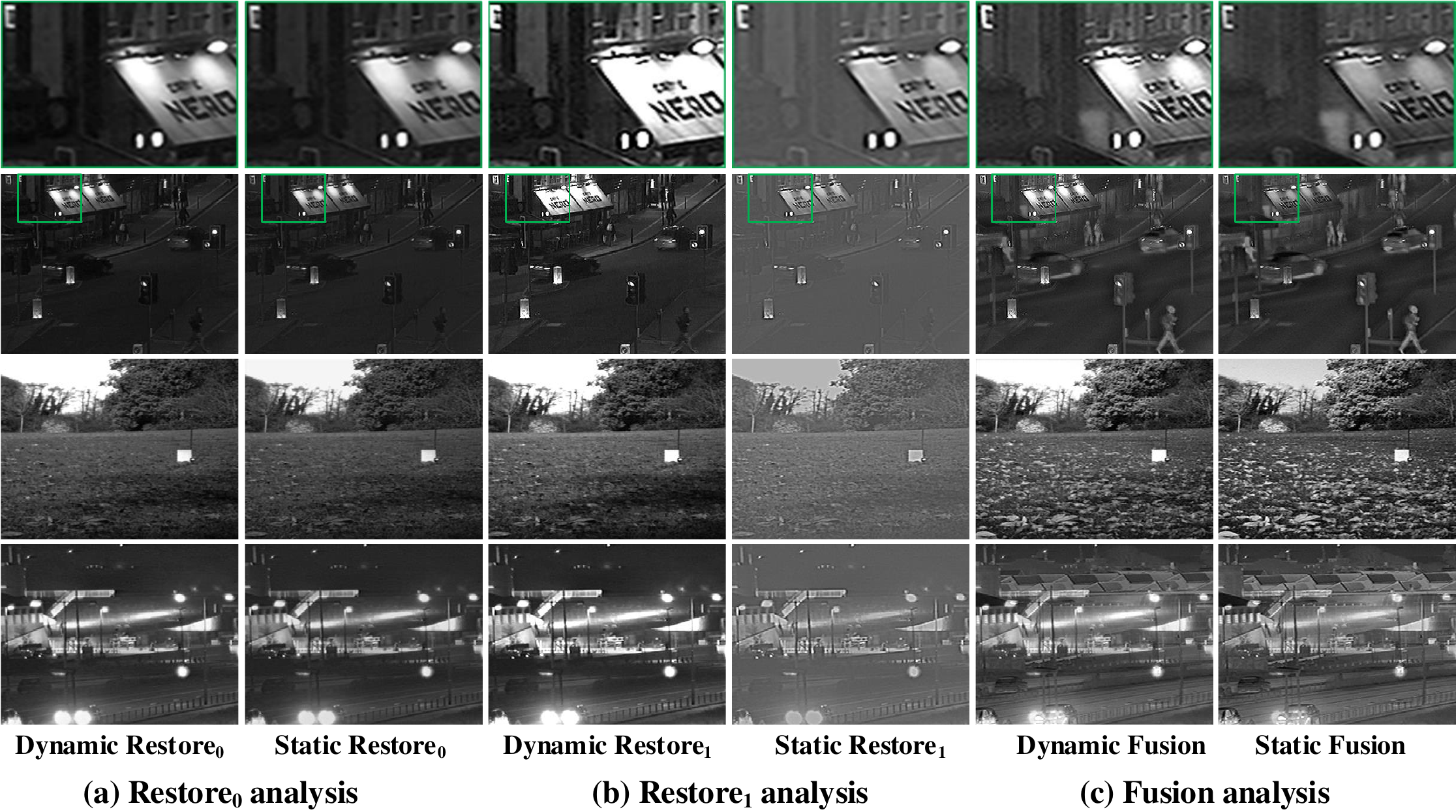}
	\caption{Static and dynamic analysis. (a) Only the image restoration loss optimized image recovery results, i.e., dynamic restore$_0$, static restore$_0$. (b) The results of the restoration are optimized by the loss of restoration and image fusion, i.e., dynamic restore$_1$, static restore$_1$. (c) Fusion analysis results, i.e., dynamic fusion, and static fusion.\label{analysis1}}
\end{figure}

From Figure \ref{analysis1}, we can draw the following three conclusions. 
(1) Compared with the joint optimization strategy of restoration loss and image fusion loss, the restoration results obtained by the simple restoration loss optimization method have better image definition, e.g., contrast, illumination, and gradient.
(2) Dynamic fusion is better than static fusion in complex environment (e.g., low-light) has better fusion quality.
(3) Under the joint optimization strategy, the static restore result is poor, and there is an obvious hazy effect. However, the result of dynamic restoration is good. (4) The static fusion model has better gradient information in the normal environment, which makes the image edge sharper. 

\subsubsection{Comparative analysis of loss and network architecture}

\begin{figure}[h]
	\centering
	\includegraphics[width=1\textwidth]{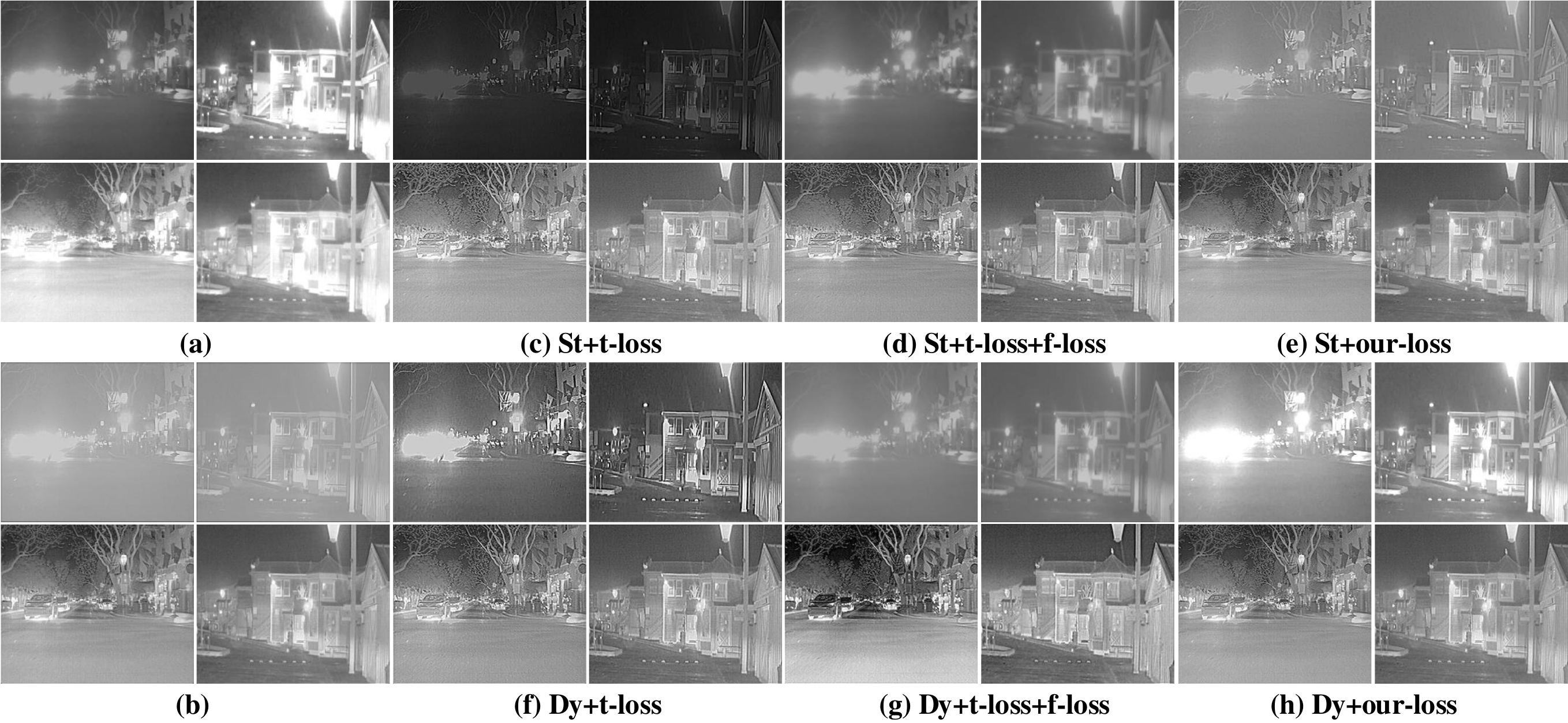}
	\caption{Loss function and network architecture analysis, $St$ denotes static network; $Dy$ denotes dynamic network; $t-loss$ denotes traditional fusion loss; $f-loss$ denotes $H_3$ fusion loss. There are four groups of experiments in static network, i.e., (a), (c), (d), and (e). There are four groups of experiments in dynamic network, i.e., (b), (f), (g), and (h).
	The first line of each group represents the image restoration experiment, and the second line represents the image fusion experiment. (a) Traditional static image fusion network, i.e., without image restoration, traditional loss, and static network. (b) Dynamic image fusion network, i.e., without image restoration, traditional loss, and dynamic network. (c) and (f): Image fusion network (i.e., static fusion, dynamic fusion, ) based on traditional similarity loss. The image restoration loss is only in the first 10 epoach optimizations. (d) and (g): Image fusion network (i.e., static fusion, dynamic fusion, ) based on traditional similarity loss. Image restoration loss and image fusion loss are jointly optimized. (e) and (h): The same loss function, different networks, i.e., static fusion network, and dynamic fusion network }
	\label{analysis2}
\end{figure}

In order to demonstrate the superiority of our loss function and network architecture, 
we compare and analyze the traditional similarity loss, traditional static fusion, and dynamic fusion networks. In this experiment, we will use the same training dataset, the same network structure, and the same training parameters. This experiment mainly explores the influence of traditional similarity loss and our loss function on image fusion, i.e., static fusion and dynamic fusion. The experimental results are shown in Figure \ref{analysis2}. The experiment includes 8 sub-experiments, i.e., 4 restoration experiments, and 4 fusion experiments.

From Figure \ref{analysis2}, we can find several phenomena. (1) The fusion results based on our fusion and restoration framework (c-f) are generally better than (a-b). This shows that our fusion framework has better performance than pure fusion framework. (2) (a, c-e) is superior to (b, f-h) in the quality of image restoration and fusion. This shows that dynamic network has more advantages in comprehensive performance than static network. (3) (d-e) and (g-h) four groups of experiments demonstrate that under the same network framework, the proposed loss can significantly improve the quality of image restoration, and (d) and (g) can solve the problem of over sharpening. (4) (c, f) compared with (e, h), the former have better gradient information. Compared with c, f has better image restoration results, and i.e., the dynamic network has better advantages.

\subsubsection{Image restore and fusion efficiency analysis}
The proposed network can complete the tasks of image restoration and image fusion at the same time.  As far as we know, the existing image fusion network cannot simultaneously take care of these two tasks. The model size and running time are shown in Table \ref{table112}. It can be seen from the experimental data that under the same conditions, this method has small model and fast running speed. In addition, our network does not need complex pre-training model and complex network structure.

\section{Discussion}
Exhaustive experiments in section \ref{setup} verify that our dynamic image fusion method is better than existing cross-modal image fusion methods in robustness. This also prove the effectiveness of our simulation of human brain dynamic cognitive mechanism. There are three potential reasons:.

\textit{1) Dynamic fusion mechanism}. Human beings can be robust in a variety of visual tasks, which is closely related to the dynamic characteristics of the human brain. This dynamic characteristic is of great significance for various visual tasks. However, the existing cross-modal image fusion methods mostly use static fusion weights or separate the fusion layer from the decoding layer, which will aggravate the difference of data distribution to a certain extent.

\textit{2) Dynamic degradation mechanism}.  Although the existing image restoration task or image super-resolution task has studied a variety of degenerative kernels, the related research is limited to the static degenerative kernels of a single image. In the field of image fusion, the research on the robustness of image fusion has not been found. This is why our research is of great significance in the field of image fusion.

\textit{3) Joint optimization loss}. Based on the loss of traditional similarity measure (i.e., the loss of similarity between the predicted image and the original image), we construct a dynamic negative sample fusion loss based on the dynamic degradation model. On the one hand, the loss function reduces the difference between the prediction result and the original data distribution by similarity loss; on the other hand, by expanding the distance between the prediction result and the dynamic degradation data distribution, the network is forced to optimize in the direction of improving the quality, rather than simply being similar to the original images.

\section{Conclusion}
In this paper, we proposed a dynamic cross-modal image restore and fusion method inspired by human brain dynamic cognitive mechanism. We mainly studied the influence of dynamic degradation model and dynamic fusion weight on image restore and fusion tasks. Our network effectively unifies image restoration tasks and image fusion tasks. The same network can complete the image restoration task and image fusion task. Our network has better feature mining ability than the existing image fusion methods. To verify the effectiveness of our DDRF-Net, we carried out a large number of comparative experiments and analytical experiments, and the experimental results show the superiority of our method.

\begin{flushleft}	
\textbf{Acknowledgment}\\
\end{flushleft}
This work was supported by the National Natural Science Foundation of China under Grants nos. 61871326, and the Shanxi Natural Science Basic Research Program under Grant no. 2018JM6116, and National Natural Science Foundation of China under Grants nos. 61231016.

\section*{References}

\bibliography{mybibfile}

\end{document}